\documentclass[10pt,twocolumn,letterpaper]{article}

\usepackage[pagenumbers]{cvpr} 

\usepackage[most]{tcolorbox}

\newtcolorbox{disclaimerbox}{
    colback=gray!10,     
    colframe=gray!40,    
    boxrule=0.5pt,       
    arc=4pt,             
    auto outer arc,
    boxsep=5pt,
    left=6pt,
    right=6pt,
    top=4pt,
    bottom=4pt,
    enhanced jigsaw
}

\definecolor{iccvblue}{rgb}{0.21,0.49,0.74}
\usepackage[pagebackref,breaklinks,colorlinks,allcolors=iccvblue]{hyperref}

\def\paperID{*****} 
\def\confName{CVPR}
\def\confYear{2026}

\title{3D Smoke Scene Reconstruction Guided by Vision Priors from Multimodal Large Language Models}

\author{
  Xinye Zheng$^{1}$ \quad
  Fei Wang$^{1,2}$ \quad
  Yiqi Nie$^{2,3}$ \quad
  Kun Li$^{4}$ \quad 
  Junjie Chen$^{1,2}$ \quad \\
  Jiaqi Zhao$^{2,3}$ \quad
  Yanyan Wei$^{1}$ \quad
  Zhiliang Wu$^{5}$ \\
  $^{1}$\,Hefei University of Technology \\
  $^{2}$\,Institute of Artificial Intelligence, Hefei Comprehensive National Science Center \\
  $^{3}$\,Anhui University \quad
  $^{4}$\,United Arab Emirates University \quad
  $^{5}$\,Nanyang Technological University \\
}

\begin{document}
\maketitle

\begin{abstract}
Reconstructing 3D scenes from smoke-degraded multi-view images is particularly difficult because smoke introduces strong scattering effects, view-dependent appearance changes, and severe degradation of cross-view consistency. To address these issues, we propose a framework that integrates visual priors with efficient 3D scene modeling. We employ Nano-Banana-Pro to enhance smoke-degraded images and provide clearer visual observations for reconstruction and develop Smoke-GS, a medium-aware 3D Gaussian Splatting framework for smoke scene reconstruction and restoration-oriented novel view synthesis. Smoke-GS models the scene using explicit 3D Gaussians and introduces a lightweight view-dependent medium branch to capture direction-dependent appearance variations caused by smoke. Our method preserves the rendering efficiency of 3D Gaussian Splatting while improving robustness to smoke-induced degradation. Results demonstrate the effectiveness of our method for generating consistent and visually clear novel views in challenging smoke environments.
\end{abstract}

\begin{disclaimerbox}
Our team (XInsight Lab) achieved \textbf{2nd place} in Track 2 of the NTIRE 2026 3D Restoration and Reconstruction Challenge, as reported on the official competition website: \url{https://www.codabench.org/competitions/13993/}.
\end{disclaimerbox}

\section{Introduction}
3D reconstruction and novel view synthesis (NVS) are foundational tasks in the computer vision community~\cite{wang2024eulermormer,wang2024frequency,wang2026task,wang2024low,guo2024benchmarking,wei2025leveraging,wu2025bvinet,wu2025drafting,wu2026dlvinet,chen2025timar, liu2026ntire, chang2026training, ge2026dual, chang2026beyond, ge2026clip, liu2026elog, fu2026smokegs, cao2026gensmoke, zhu2026naka, guo2026reliability, chen2026dehaze}, and have great potential in fields such as embodied intelligence, robotics, and autonomous driving. Accurate 3D geometry reconstruction and view-consistent rendering directly determine the reliability and generalization capability of high-level perception, planning, and interaction systems.

The NTIRE 2026 3D Restoration and Reconstruction Challenge~\cite{liu2026ntire3d} focuses on the visual degradation and geometric inconsistency caused by physical degradations in real-world scenarios. Track 2 is dedicated to 3D smoke restoration, providing an authoritative platform for evaluating and advancing 3D reconstruction methods in scattering media. The challenge requires participating methods to construct efficient and robust 3D reconstruction solutions from diverse real-world multi-view image sequences captured under smoke interference, with a particular focus on addressing the multi-view inconsistency and geometric degradation caused by scattering media.

As a typical scattering medium, smoke introduces view-dependent optical attenuation, non-uniform contrast reduction, and dense structural occlusions, which severely compromise multi-view geometric consistency and make it difficult for conventional methods to effectively recover scene geometry and appearance. The challenge demands approaches capable of building efficient and robust 3D reconstruction pipelines from real-world multi-view image sequences captured in the presence of smoke interference.

To address the above challenges, we propose a novel 3D reconstruction method, Smoke-GS. This approach employs Nano-Banana-Pro to enhance the input images, mitigating visual degradation caused by smoke, and then performs 3D reconstruction and novel-view rendering based on the enhanced images.

\section{Related Work}

\subsection{Single Image Dehazing}

\begin{figure*}[!t]
\centering
\includegraphics[width=\textwidth]{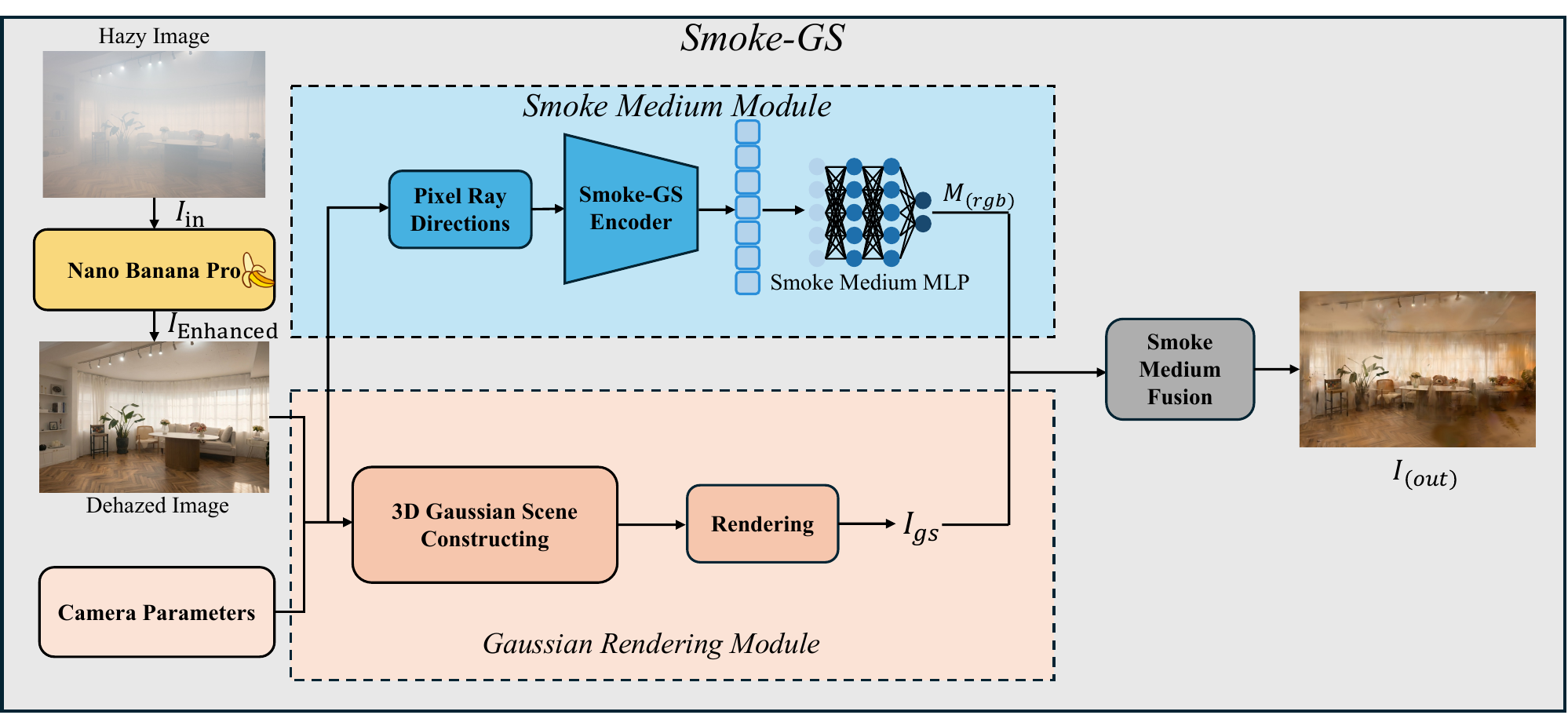}
\caption{Overview of our Smoke-GS method for hazy image restoration and 3D reconstruction. The pipeline begins with a hazy image that is enhanced using Nano Banana Pro. The enhanced image, along with the dehazed image and camera parameters, is used to generate the Smoke Medium Module. This module encodes pixel ray directions through the Smoke-GS Encoder, which feeds into a Smoke Medium MLP to predict medium parameters. These medium terms are fused with the 3D Gaussian rendering to produce the final output image.}
\label{fig:arch}
\end{figure*}

Similar to other image restoration tasks~\cite{wei2021deraincyclegan, chen2022snowformer, chen2023sparse, jin2022unsupervised}, single-image dehazing aims to recover clear images from hazy observations. Learning-based methods~\cite{cai2016dehazenet, cui2023focal, dong2020multi} have come to dominate this field. For instance, MSBDN~\cite{dong2020multi} leverages boosting and error feedback for iterative refinement. Transformer architectures and diffusion models~\cite{chen2024teaching, qiu2023mb, song2023vision, qian2025physdiff} have further improved dehazing performance. Despite these advances, recent research has increasingly shifted toward real-world haze, which typically requires specialized model designs and training strategies yet still faces performance limitations under diverse and complex conditions.

\subsection{Image Generation}
Text-to-image diffusion models~\cite{hao2023optimizing, balaji2022ediff, xu2023imagereward,qian2024cluster,qian2024joint} have become central to generating high-quality images from prompts. Among them, Stable Diffusion~\cite{rombach2022high} excels by performing diffusion in latent space. Recent works have further extended its capabilities: DreamBooth~\cite{ruiz2023dreambooth} and Custom-Diffusion~\cite{kumari2023multi} fine-tune models for specific content, while T2I-Adapter~\cite{mou2024t2i} and ControlNet~\cite{zhang2023adding} improve the precision of image control. For other tasks, InstructPix2Pix~\cite{brooks2023instructpix2pix} adapts Stable Diffusion for image editing, and CycleGAN-Turbo~\cite{parmar2024one} integrates GANs for diverse image translation. GenDeg~\cite{rajagopalan2025gendeg} leverages InstructPix2Pix to construct large-scale degradation datasets.

\subsection{3D Reconstruction under Scattering Media}
In fog or smoke environments, scattering and absorption reduce transmittance and introduce additive path radiance. Therefore, reconstruction methods often incorporate radiative-transfer terms to decouple direct scene radiance from scattered components under multi-view constraints. A representative approach integrates physically grounded transmittance and in-scattering into neural rendering to achieve joint reconstruction and dehazing~\cite{ramazzina2023scatternerf}. Subsequent work~\cite{li2023dehazing, chen2024dehazenerf, zhang2025decoupling,zhang2026tg4mm} further improves model robustness by introducing additional physical priors, as well as by transferring the same image formation model to 3D Gaussian Splatting pipelines~\cite{ma2025dehazegs}.

On the one hand, physics-based scattering models have been extended to Gaussian Splatting representations, thereby improving reconstruction efficiency through media-aware rendering formulations~\cite{li2025watersplatting, yang2025seasplat,wang2026multimodal,wang2026xinsight}. On the other hand, incorporating stronger priors to separate media effects from scene appearance~\cite{tang2024neural, gough2025aquanerf, wu2025plenodium, guo2025neuropump}, or constructing unified media-interaction models that handle scattering and illumination coupling within a single framework~\cite{liu2025i2nerf}, has become an important direction for improving 3D reconstruction robustness under smoke environments.

\section{Methodology}

As shown in Figure~\ref{fig:arch}, we propose Smoke-GS, a novel view restoration framework for smoke-degraded scenes. Built on 3D Gaussian Splatting, Smoke-GS introduces a lightweight, view-dependent medium branch to model direction-dependent appearance variations caused by smoke. By combining an explicit 3D scene representation with medium-aware appearance compensation, our proposed method can synthesize geometrically consistent and visually clear images from arbitrary target viewpoints.

\subsection{Problem Definition}

Given a set of multi-view smoke-degraded images
\begin{equation}
    \{I_1, I_2, \dots, I_N\},
\end{equation}
where each image $I_i \in \mathbb{R}^{H \times W \times 3}$, together with the corresponding camera parameters
\begin{equation}
    \{P_1, P_2, \dots, P_N\},
\end{equation}
where $P_i$ includes both camera intrinsics $(f_x, f_y, c_x, c_y)$ and extrinsics, our goal is to learn a 3D scene representation that supports geometrically consistent and visually realistic rendering under arbitrary viewpoints. Given a target camera pose $P_{\text{test}}$, Smoke-GS predicts a restored image $\hat{I}$ that approximates the clear appearance of the underlying scene.

\subsection{3D Gaussian Scene Representation}
\begin{table*}[t]
\caption{Results on the Track 2 final leaderboard. PSNR and SSIM are reported for novel view synthesis under smoke-degraded scenes.}
\label{tab:track2_results}
\centering
\renewcommand{\arraystretch}{1.1}
\begin{tabular}{cccccc}
\hline
Rank & Team name & User Name & PSNR & SSIM & Affiliation \\
\hline
1 & plbbl & pilibaobaolong & 20.2061 & 0.7263 & Hangzhou Dianzi University \\
\textbf{2} & \textbf{XInsight Lab} & \textbf{shaneyale} & \textbf{18.6681} & \textbf{0.6909} & \textbf{Hefei University of Technology} \\
3 & Hunan Duo & hqking & 18.3816 & 0.6583 & Huazhong University of Science and Technology \\
4 & AIIA\_LAB & guyang\_2 & 17.5486 & 0.6943 & Harbin Institute of Technology \\
5 & 3ddd & 2905523 & 16.6687 & 0.6643 & Insta360 research \\
6 & Diouj. El & Diouj. El & 15.3625 & 0.6574 & VNU University of Science \\
7 & windrise & windrise & 15.2174 & 0.6657 & University of Science and Technology \\
8 & zzz & clbqyxtj & 14.8923 & 0.6103 & Xidian University, Intelligent Perception and... \\
9 & dlmath\_vision & dlmath\_vision & 14.4005 & 0.6341 & Korea University, Department of Mathematics \\
10 & EE-GS & jumbooling & 14.3759 & 0.6569 & Xidian University \\
11 & Helicopter & beatlakers & 13.8623 & 0.5859 & Pengcheng Laboratory \\
12 & MonoSmokeGS & linzhe & 13.6479 & 0.6251 & N/A \\
13 & AAA & tjuzhan & 13.5287 & 0.6181 & Tianjin University \\
14 & ffffyb & ffffyb & 12.0852 & 0.5335 & Hainan University \\
\hline
\end{tabular}
\end{table*}

Following 3D Gaussian Splatting, we represent the scene using a set of learnable 3D Gaussians $\{G_1, G_2, \dots, G_M\}$. 
Each Gaussian $G_j$ is parameterized as
\begin{equation}
    G_j = \{\mu_j, q_j, s_j, \alpha_j, SH_j\},
\end{equation}
where $\mu_j \in \mathbb{R}^3$ denotes the 3D position, $q_j \in \mathbb{R}^4$ is the rotation quaternion, $s_j \in \mathbb{R}^3$ represents the logarithmic scale along the three principal axes, $\alpha_j \in \mathbb{R}$ is the opacity, and $SH_j \in \mathbb{R}^{K \times 3}$ denotes the spherical harmonics coefficients for view-dependent color modeling.

We initialize the scene with $M=100{,}000$ Gaussians. The maximum degree of spherical harmonics is set to $D=3$, yielding
\begin{equation}
    K=(D+1)^2=16.
\end{equation}
To stabilize optimization in the early stage of training, the spherical harmonics degree is gradually increased from $D=0$ to $D=3$.

\subsection{Gaussian Scene Branch}

Given a target camera pose matrix $T_{\text{cam}} \in \mathbb{R}^{4 \times 4}$, image resolution $(H,W)$, and intrinsic matrix $K \in \mathbb{R}^{3 \times 3}$, the Gaussian Scene Branch first converts the camera pose into a view matrix $V = T_{\text{cam}}^{-1}$, and multiplies the second and third rows by $-1$ to match the rendering coordinate convention.

The differentiable rasterize is applied to render the basic scene appearance:
\begin{equation}
    I_{\text{gs}} = \mathrm{Rasterize}\left(
    \{\mu_j, q_j, \exp(s_j), \sigma(\alpha_j), SH_j\}, V, K, H, W
    \right),
\end{equation}
where $\sigma(\cdot)$ denotes the Sigmoid activation function. Here $\exp(s_j)$ converts the logarithmic scale to the actual Gaussian scale, and $\sigma(\alpha_j)$ constrains the opacity to the range $(0,1)$. The output $I_{\text{gs}} \in \mathbb{R}^{H \times W \times 3}$ serves as the base rendering of the scene.

\subsection{View-Dependent Medium Branch}

Although 3DGS can model scene structure and view-dependent appearance effectively, smoke introduces additional direction-dependent visual effects, such as color shift, contrast attenuation, and scattering-related appearance variation. To address this issue, we design a lightweight View-Dependent Medium Branch that predicts a medium-aware correction term for each pixel according to its viewing direction.

For each pixel $(u,v)$, we first map it into the normalized camera coordinate system:
\begin{equation}
    x = \frac{u-c_x}{f_x}, \qquad
    y = \frac{v-c_y}{f_y}, \qquad
    z = 1.
\end{equation}
The corresponding unit ray direction is then computed as
\begin{equation}
    \mathbf{d} = \frac{(x,y,z)}{\|(x,y,z)\|}.
\end{equation}
Let $R \in \mathbb{R}^{3 \times 3}$ denote the camera rotation matrix. The ray direction in the world coordinate system is given by
\begin{equation}
    \mathbf{d}_{\text{world}} = R^\top \mathbf{d}.
\end{equation}

To better capture directional variation, we encode the world-space ray direction using 4th-order spherical harmonics:
\begin{equation}
    \mathbf{f}_{\text{dir}} = SH(\mathbf{d}_{\text{world}}) \in \mathbb{R}^{25}.
\end{equation}
Spherical harmonics provide an efficient representation of smoothly varying directional signals and are well-suited to modeling the continuous appearance changes caused by smoke across different viewing directions.

The encoded directional feature is then fed into a lightweight two-layer multilayer perceptron:
\begin{equation}
    \mathbf{h} = \sigma(W_1 \mathbf{f}_{\text{dir}} + \mathbf{b}_1),
\end{equation}
\begin{equation}
    \mathbf{o} = W_2 \mathbf{h} + \mathbf{b}_2,
\end{equation}
where $\sigma$ denotes the Sigmoid activation function, and the hidden dimension is set to $128$.

The network outputs $9$ channels, which are decomposed into three groups:
\begin{equation}
    \mathbf{o} = \{\text{medium}_{rgb}, \text{medium}_{bs}, \text{medium}_{attn}\},
\end{equation}
where $\text{medium}_{rgb} \in \mathbb{R}^3$ denotes the medium-induced color correction term and is activated by Sigmoid, $\text{medium}_{bs} \in \mathbb{R}^3$ denotes the backward scattering component and is activated by Softplus, and $\text{medium}_{attn} \in \mathbb{R}^3$ denotes the attenuation component and is also activated by Softplus.

Currently, we adopt a simple yet effective fusion strategy and only use $\text{medium}_{rgb}$ as the view-dependent correction term:
\begin{equation}
    I_{\text{output}} = I_{\text{gs}} + 0.2 \cdot \text{medium}_{rgb}.
\end{equation}

\section{Experiment}

\subsection{Experiment Setup}
\textbf{Datasets.} We conduct the experiments on the RealX3D dataset~\cite{liu2025realx3d}, a real-capture benchmark for multi-view visual restoration and 3D reconstruction under physical degradations. This work focuses on Track 2 of the NTIRE 2026 Challenge, 3D Smoke Restoration, using the smoke-scattering subset of RealX3D. This subset contains five smoke scenes, each providing multi-view training images, corresponding camera poses, and NVS target poses. The dataset is divided into a validation set of 1 scene, a development set of 4 scenes, and a test set of 3 scenes. For the development and test sets, only NVS target poses are provided without reference images.

\textbf{Evaluation Metrics.}
We adopt Peak Signal-to-Noise Ratio (PSNR) and Structural Similarity Index (SSIM) as the evaluation metrics for photometric fidelity. PSNR measures pixel-level reconstruction error and is sensitive to absolute intensity differences. SSIM evaluates image similarity based on luminance, contrast, and structure, aligning more closely with human visual perception. For the development and test sets, we submit the rendered novel NVS images to the challenge server for online evaluation. The server computes the metrics using undisclosed ground-truth reference images and releases the results.

\textbf{Implementation Details.} Our method is trained on a single NVIDIA A800 GPU, with a separate model independently optimized for each scene. 
The training input consists of multi-view RGB images and their corresponding camera parameters. For each scene, we initialize 100,000 3D Gaussians and enable the Smoke Medium MLP to predict \textit{medium\_rgb}, \textit{medium\_bs}, and \textit{medium\_attn}. The total number of training steps is set to 150,000, and the Adam optimizer is used with different learning rates for various parameter groups. 
The initial learning rates for position, rotation, scale, and opacity are 1.6e-4, 1e-3, 5e-3, and 5e-2, respectively, and the learning rate for the MLP is 1e-3. The loss function is a weighted combination of L1 and SSIM, defined as \( L = (1 - \lambda)L1 + \lambda(1 - \text{SSIM}) \), with \( \lambda = 0.2 \).

\subsection{Results}

The quantitative comparison is presented in Table~\ref{tab:track2_results}. Our submission ranks \textbf{2nd} on the final leaderboard, achieving a PSNR of \textbf{18.6681} and an SSIM of \textbf{0.6909}. This result verifies the effectiveness of the proposed Smoke-GS framework on real-world smoke-degraded 3D reconstruction and novel view synthesis. Our Smoke-GS outperforms the 3rd-ranked submission by \textbf{0.2865} dB in PSNR and \textbf{0.0326} in SSIM, indicating stronger reconstruction fidelity and better structural preservation. Compared with most participating methods, Smoke-GS shows a clear advantage in overall quality.

\section{Conclusion}
We presented Smoke-GS, our solution for Track 2 of the NTIRE 2026 3D Restoration and Reconstruction Challenge. By combining visual enhancement from Nano-Banana-Pro with a medium-aware 3D Gaussian Splatting framework, our proposed method improves restoration-oriented NVS in smoke-degraded scenes. The final results show that our method achieves strong performance on the official benchmark and ranks \textbf{2nd} in the challenge leaderboard, demonstrating the effectiveness of integrating visual priors with explicit 3D scene representation for smoke scene reconstruction.

{
    \small
    \bibliographystyle{ieeenat_fullname}
    \bibliography{main}

@String(AAAI = {AAAI})

@article{liu2025realx3d,
  title={RealX3D: A Physically-Degraded 3D Benchmark for Multi-view Visual Restoration and Reconstruction},
  author={Liu, Shuhong and Bao, Chenyu and Cui, Ziteng and Liu, Yun and Chu, Xuangeng and Gu, Lin and Conde, Marcos V and Umagami, Ryo and Hashimoto, Tomohiro and Hu, Zijian and others},
  journal={arXiv preprint arXiv:2512.23437},
  year={2025}
}

@inproceedings{liu2025i2nerf,
  title={I2-NeRF: Learning Neural Radiance Fields Under Physically-Grounded Media Interactions},
  author={Liu, Shuhong and Gu, Lin and Cui, Ziteng and Chu, Xuangeng and Harada, Tatsuya},
  booktitle={Advances in Neural Information Processing Systems (NeurIPS)},
  year={2025},
}

@article{liu2026ntire3d,
  title={NTIRE 2026 3D Restoration and Reconstruction in Real-world Adverse Conditions: RealX3D Challenge Results},
  author={Liu, Shuhong and Bao, Chenyu and Cui, Ziteng and others},
  journal={arXiv preprint arXiv:2604.04135},
  year={2026}
}

@inproceedings{li2025watersplatting,
  title={Watersplatting: Fast underwater 3d scene reconstruction using gaussian splatting},
  author={Li, Huapeng and Song, Wenxuan and Xu, Tianao and Elsig, Alexandre and Kulhanek, Jonas},
  booktitle={International Conference on 3D Vision},
  pages={969--978},
  year={2025},
  organization={IEEE}
}

@inproceedings{yang2025seasplat,
  title={Seasplat: Representing underwater scenes with 3d gaussian splatting and a physically grounded image formation model},
  author={Yang, Daniel and Leonard, John J and Girdhar, Yogesh},
  booktitle={IEEE International Conference on Robotics and Automation},
  pages={7632--7638},
  year={2025},
  organization={IEEE}
}

@inproceedings{wang2024eulermormer,
  title={Eulermormer: Robust eulerian motion magnification via dynamic filtering within transformer},
  author={Wang, Fei and Guo, Dan and Li, Kun and Wang, Meng},
  booktitle={Proceedings of the AAAI Conference on Artificial Intelligence},
  volume={38},
  number={6},
  pages={5345--5353},
  year={2024}
}

@inproceedings{wang2024frequency,
  title={Frequency decoupling for motion magnification via multi-level isomorphic architecture},
  author={Wang, Fei and Guo, Dan and Li, Kun and Zhong, Zhun and Wang, Meng},
  booktitle={Proceedings of the IEEE/CVF Conference on Computer Vision and Pattern Recognition},
  pages={18984--18994},
  year={2024}
}

@article{wang2026task,
  title={Task-generalized adaptive cross-domain learning for multimodal image fusion},
  author={Wang, Mengyu and Liu, Zhenyu and Li, Kun and Wang, Yu and Wang, Yuwei and Wei, Yanyan and Wang, Fei},
  journal={IEEE Transactions on Multimedia},
  year={2026},
  publisher={IEEE}
}

@article{wang2024low,
  title={Low-light wheat image enhancement using an explicit inter-channel sparse transformer},
  author={Wang, Yu and Wang, Fei and Li, Kun and Feng, Xuping and Hou, Wenhui and Liu, Lu and Chen, Liqing and He, Yong and Wang, Yuwei},
  journal={Computers and Electronics in Agriculture},
  volume={224},
  pages={109169},
  year={2024},
  publisher={Elsevier}
}

@inproceedings{wu2025bvinet,
  title={BVINet: Unlocking blind video inpainting with zero annotations},
  author={Wu, Zhiliang and Chen, Kerui and Li, Kun and Fan, Hehe and Yang, Yi},
  booktitle={Proceedings of the IEEE/CVF International Conference on Computer Vision},
  pages={14017--14027},
  year={2025}
}

@inproceedings{wu2026dlvinet,
  title={DLVINet: Advancing dual-lens video inpainting beyond parallax constraints},
  author={Wu, Zhiliang and Li, Kun and Xu, Yunqiu and Fan, Hehe and Yang, Yi},
  booktitle={Proceedings of the AAAI Conference on Artificial Intelligence},
  volume={40},
  number={13},
  pages={10888--10896},
  year={2026}
}

@inproceedings{wu2025drafting,
  title={Drafting and revision: advancing high-fidelity video inpainting},
  author={Wu, Zhiliang and Li, Kun and Fan, Hehe and Yang, Yi},
  booktitle={Proceedings of the Thirty-Fourth International Joint Conference on Artificial Intelligence},
  pages={2063--2071},
  year={2025}
}

@inproceedings{ramazzina2023scatternerf,
  title={Scatternerf: Seeing through fog with physically-based inverse neural rendering},
  author={Ramazzina, Andrea and Bijelic, Mario and Walz, Stefanie and Sanvito, Alessandro and Scheuble, Dominik and Heide, Felix},
  booktitle={Proceedings of the IEEE/CVF international conference on computer vision},
  pages={17957--17968},
  year={2023}
}

@article{li2023dehazing,
  title={Dehazing-NeRF: neural radiance fields from hazy images},
  author={Li, Tian and Li, LU and Wang, Wei and Feng, Zhangchi},
  journal={arXiv preprint arXiv:2304.11448},
  year={2023}
}

@inproceedings{chen2024dehazenerf,
  title={DehazeNeRF: Multi-image haze removal and 3D shape reconstruction using neural radiance fields},
  author={Chen, Wei-Ting and Yifan, Wang and Kuo, Sy-Yen and Wetzstein, Gordon},
  booktitle={2024 International Conference on 3D Vision (3DV)},
  pages={247--256},
  year={2024},
  organization={IEEE}
}

@inproceedings{zhang2025decoupling,
  title={Decoupling Scattering: Pseudo-Label Guided NeRF for Scenes with Scattering Media},
  author={Zhang, Mingyang and Zhang, Junkang and Fang, Faming and Zhang, Guixu},
  booktitle={Proceedings of the AAAI Conference on Artificial Intelligence},
  volume={39},
  number={10},
  pages={10031--10039},
  year={2025}
}

@article{ma2025dehazegs,
  title={DehazeGS: 3D Gaussian splatting for multi-image haze removal},
  author={Ma, Chenjun and Zhao, Jieyu and Chen, Jian},
  journal={IEEE Signal Processing Letters},
  volume={32},
  pages={736--740},
  year={2025},
  publisher={IEEE}
}

@inproceedings{tang2024neural,
  title={Neural underwater scene representation},
  author={Tang, Yunkai and Zhu, Chengxuan and Wan, Renjie and Xu, Chao and Shi, Boxin},
  booktitle={Proceedings of the IEEE/CVF Conference on Computer Vision and Pattern Recognition},
  pages={11780--11789},
  year={2024}
}

@inproceedings{gough2025aquanerf,
  title={AquaNeRF: Neural radiance fields in underwater media with distractor removal},
  author={Gough, Luca and Azzarelli, Adrian and Zhang, Fan and Anantrasirichai, Nantheera},
  booktitle={2025 IEEE International Symposium on Circuits and Systems (ISCAS)},
  pages={1--5},
  year={2025},
  organization={IEEE}
}

@article{wu2025plenodium,
  title={Plenodium: UnderWater 3D Scene Reconstruction with Plenoptic Medium Representation},
  author={Wu, Changguanng and Dong, Jiangxin and Li, Chengjian and Tang, Jinhui},
  journal={arXiv preprint arXiv:2505.21258},
  year={2025}
}

@inproceedings{guo2025neuropump,
  title={Neuropump: Simultaneous geometric and color rectification for underwater images},
  author={Guo, Yue and Liao, Haoxiang and Ling, Haibin and Huang, Bingyao},
  booktitle={Proceedings of the 33rd ACM International Conference on Multimedia},
  pages={422--431},
  year={2025}
}

@article{chen2022snowformer,
  title={SnowFormer: Context interaction transformer with scale-awareness for single image desnowing},
  author={Chen, Sixiang and Ye, Tian and Liu, Yun and Chen, Erkang},
  journal={arXiv preprint arXiv:2208.09703},
  year={2022}
}

@inproceedings{chen2023sparse,
  title={Sparse sampling transformer with uncertainty-driven ranking for unified removal of raindrops and rain streaks},
  author={Chen, Sixiang and Ye, Tian and Bai, Jinbin and Chen, Erkang and Shi, Jun and Zhu, Lei},
  booktitle={Proceedings of the IEEE/CVF international conference on computer vision},
  pages={13106--13117},
  year={2023}
}

@inproceedings{jin2022unsupervised,
  title={Unsupervised night image enhancement: When layer decomposition meets light-effects suppression},
  author={Jin, Yeying and Yang, Wenhan and Tan, Robby T},
  booktitle={European conference on computer vision},
  pages={404--421},
  year={2022},
  organization={Springer}
}

@article{cai2016dehazenet,
  title={Dehazenet: An end-to-end system for single image haze removal},
  author={Cai, Bolun and Xu, Xiangmin and Jia, Kui and Qing, Chunmei and Tao, Dacheng},
  journal={IEEE transactions on image processing},
  volume={25},
  number={11},
  pages={5187--5198},
  year={2016},
  publisher={IEEE}
}

@inproceedings{cui2023focal,
  title={Focal network for image restoration},
  author={Cui, Yuning and Ren, Wenqi and Cao, Xiaochun and Knoll, Alois},
  booktitle={Proceedings of the IEEE/CVF international conference on computer vision},
  pages={13001--13011},
  year={2023}
}

@inproceedings{dong2020multi,
  title={Multi-scale boosted dehazing network with dense feature fusion},
  author={Dong, Hang and Pan, Jinshan and Xiang, Lei and Hu, Zhe and Zhang, Xinyi and Wang, Fei and Yang, Ming-Hsuan},
  booktitle={Proceedings of the IEEE/CVF conference on computer vision and pattern recognition},
  pages={2157--2167},
  year={2020}
}

@inproceedings{chen2024teaching,
  title={Teaching tailored to talent: Adverse weather restoration via prompt pool and depth-anything constraint},
  author={Chen, Sixiang and Ye, Tian and Zhang, Kai and Xing, Zhaohu and Lin, Yunlong and Zhu, Lei},
  booktitle={European Conference on Computer Vision},
  pages={95--115},
  year={2024},
  organization={Springer}
}

@inproceedings{qiu2023mb,
  title={Mb-taylorformer: Multi-branch efficient transformer expanded by taylor formula for image dehazing},
  author={Qiu, Yuwei and Zhang, Kaihao and Wang, Chenxi and Luo, Wenhan and Li, Hongdong and Jin, Zhi},
  booktitle={Proceedings of the IEEE/CVF international conference on computer vision},
  pages={12802--12813},
  year={2023}
}

@article{song2023vision,
  title={Vision transformers for single image dehazing},
  author={Song, Yuda and He, Zhuqing and Qian, Hui and Du, Xin},
  journal={IEEE Transactions on Image Processing},
  volume={32},
  pages={1927--1941},
  year={2023},
  publisher={IEEE}
}

@article{guo2024benchmarking,
  title={Benchmarking Micro-action Recognition: Dataset, Methods, and Applications},
  author={Guo, Dan and Li, Kun and Hu, Bin and Zhang, Yan and Wang, Meng},
  journal={IEEE Transactions on Circuits and Systems for Video Technology},
  year={2024},
  volume={34},
  number={7},
  pages={6238-6252},
}

@article{wei2025leveraging,
  title={Leveraging vision-language prompts for real-world image restoration and enhancement},
  author={Wei, Yanyan and Zhang, Yilin and Li, Kun and Wang, Fei and Tang, Shengeng and Zhang, Zhao},
  journal={Computer Vision and Image Understanding},
  volume={250},
  pages={104222},
  year={2025},
  publisher={Elsevier}
}

@article{hao2023optimizing,
  title={Optimizing prompts for text-to-image generation},
  author={Hao, Yaru and Chi, Zewen and Dong, Li and Wei, Furu},
  journal={Advances in Neural Information Processing Systems},
  volume={36},
  pages={66923--66939},
  year={2023}
}

@article{xu2023imagereward,
  title={Imagereward: Learning and evaluating human preferences for text-to-image generation},
  author={Xu, Jiazheng and Liu, Xiao and Wu, Yuchen and Tong, Yuxuan and Li, Qinkai and Ding, Ming and Tang, Jie and Dong, Yuxiao},
  journal={Advances in Neural Information Processing Systems},
  volume={36},
  pages={15903--15935},
  year={2023}
}

@article{balaji2022ediff,
  title={ediff-i: Text-to-image diffusion models with an ensemble of expert denoisers},
  author={Balaji, Yogesh and Nah, Seungjun and others},
  journal={arXiv preprint arXiv:2211.01324},
  year={2022}
}

@inproceedings{rombach2022high,
  title={High-resolution image synthesis with latent diffusion models},
  author={Rombach, Robin and Blattmann, Andreas and Lorenz, Dominik and Esser, Patrick and Ommer, Bj{\"o}rn},
  booktitle={Proceedings of the IEEE/CVF conference on computer vision and pattern recognition},
  pages={10684--10695},
  year={2022}
}

@inproceedings{ruiz2023dreambooth,
  title={Dreambooth: Fine tuning text-to-image diffusion models for subject-driven generation},
  author={Ruiz, Nataniel and Li, Yuanzhen and Jampani, Varun and Pritch, Yael and Rubinstein, Michael and Aberman, Kfir},
  booktitle={Proceedings of the IEEE/CVF conference on computer vision and pattern recognition},
  pages={22500--22510},
  year={2023}
}

@inproceedings{kumari2023multi,
  title={Multi-concept customization of text-to-image diffusion},
  author={Kumari, Nupur and Zhang, Bingliang and Zhang, Richard and Shechtman, Eli and Zhu, Jun-Yan},
  booktitle={Proceedings of the IEEE/CVF conference on computer vision and pattern recognition},
  pages={1931--1941},
  year={2023}
}

@inproceedings{mou2024t2i,
  title={T2i-adapter: Learning adapters to dig out more controllable ability for text-to-image diffusion models},
  author={Mou, Chong and Wang, Xintao and Xie, Liangbin and Wu, Yanze and Zhang, Jian and Qi, Zhongang and Shan, Ying},
  booktitle={Proceedings of the AAAI conference on artificial intelligence},
  volume={38},
  number={5},
  pages={4296--4304},
  year={2024}
}

@inproceedings{zhang2023adding,
  title={Adding conditional control to text-to-image diffusion models},
  author={Zhang, Lvmin and Rao, Anyi and Agrawala, Maneesh},
  booktitle={Proceedings of the IEEE/CVF international conference on computer vision},
  pages={3836--3847},
  year={2023}
}

@inproceedings{brooks2023instructpix2pix,
  title={Instructpix2pix: Learning to follow image editing instructions},
  author={Brooks, Tim and Holynski, Aleksander and Efros, Alexei A},
  booktitle={Proceedings of the IEEE/CVF conference on computer vision and pattern recognition},
  pages={18392--18402},
  year={2023}
}

@article{parmar2024one,
  title={One-step image translation with text-to-image models},
  author={Parmar, Gaurav and Park, Taesung and Narasimhan, Srinivasa and Zhu, Jun-Yan},
  journal={arXiv preprint arXiv:2403.12036},
  year={2024}
}

@inproceedings{rajagopalan2025gendeg,
  title={Gendeg: Diffusion-based degradation synthesis for generalizable all-in-one image restoration},
  author={Rajagopalan, Sudarshan and Nair, Nithin Gopalakrishnan and Paranjape, Jay N and Patel, Vishal M},
  booktitle={Proceedings of the IEEE/CVF Conference on Computer Vision and Pattern Recognition},
  pages={28144--28154},
  year={2025}
}

@article{wang2026multimodal,
  title={Multimodal Protein Language Models for Enzyme Kinetic Parameters: From Substrate Recognition to Conformational Adaptation},
  author={Wang, Fei and Zheng, Xinye and Li, Kun and Wei, Yanyan and Liu, Yuxin and Hu, Ganpeng and Bao, Tong and Yang, Jingwen},
  journal={arXiv preprint arXiv:2603.12845},
  year={2026}
}

@article{wang2026xinsight,
  title={XInsight: Integrative Stage-Consistent Psychological Counseling Support Agents for Digital Well-Being},
  author={Wang, Fei and Yang, Jiangnan and Chen, Junjie and Liu, Yuxin and Li, Kun and Wei, Yanyan and Guo, Dan and Wang, Meng},
  journal={arXiv preprint arXiv:2603.06583},
  year={2026}
}

@article{zhang2026tg4mm,
  title={TG4MM: Time-Varying Gaussian Splatting for 3D Motion Magnification},
  author={Zhang, Zheng and Guo, Jiabao and Wang, Fei and Huang, Jinyang and Liu, Zhi and Guo, Dan},
  journal={IEEE Transactions on Circuits and Systems for Video Technology},
  year={2026}
}

@inproceedings{qian2024cluster,
  title={Cluster-Phys: Facial Clues Clustering Towards Efficient Remote Physiological Measurement},
  author={Qian, Wei and Li, Kun and Guo, Dan and Hu, Bin and Wang, Meng},
  booktitle={Proceedings of the 32nd ACM International Conference on Multimedia},
  pages={330--339},
  year={2024}
}

@article{qian2025physdiff,
  title={PhysDiff: Physiology-based Dynamicity Disentangled Diffusion Model for Remote Physiological Measurement},
  author={Qian, Wei and Su, Gaoji and Guo, Dan and Zhou, Jinxing and Li, Xiaobai and Hu, Bin and Tang, Shengeng and Wang, Meng},
  journal={Proceedings of the AAAI Conference on Artificial Intelligence (AAAI)},
  year={2025}
}

@article{qian2024joint,
  title={Joint Spatial-Temporal Modeling and Contrastive Learning for Self-supervised Heart Rate Measurement},
  author={Qian, Wei and Li, Qi and Li, Kun and Wang, Xinke and Sun, Xiao and Wang, Meng and Guo, Dan},
  journal={arXiv preprint arXiv:2406.04942},
  year={2024}
}

@article{chen2025timar,
  title={Towards Seamless Interaction: Causal Turn-Level Modeling of Interactive 3D Conversational Head Dynamics},
  author={Chen, Junjie and Wang, Fei and Hunag, Zhihao and Zhou, Qing and Li, Kun and Guo, Dan and Zhang, Linfeng and Yang, Xun},
  journal={arXiv preprint arXiv:2512.15340},
  year={2025}
}

@article{wei2021deraincyclegan,
  title={Deraincyclegan: Rain attentive cyclegan for single image deraining and rainmaking},
  author={Wei, Yanyan and Zhang, Zhao and Wang, Yang and Xu, Mingliang and Yang, Yi and Yan, Shuicheng and Wang, Meng},
  journal={IEEE Transactions on Image Processing},
  volume={30},
  pages={4788--4801},
  year={2021},
  publisher={IEEE}
}

@article{liu2026ntire,
 	title={NTIRE 2026 3D Restoration and Reconstruction in Real-world Adverse Conditions: RealX3D Challenge Results},
 	author={Liu, Shuhong and Bao, Chenyu and Cui, Ziteng and Chu, Xuangeng and Ren, Bin and Gu, Lin and Chen, Xiang and Li, Mingrui and Ma, Long and Conde, Marcos V and others},
 	journal={arXiv preprint arXiv:2604.04135},
 	year={2026}
}

@article{chang2026training,
 	title={Training-Free Model Ensemble for Single-Image Super-Resolution via Strong-Branch Compensation},
 	author={Chang, Gengjia and Ge, Xining and Yuan, Weijun and Li, Zhan and Song, Qiurong and Zhu, Luen and Liu, Shuhong},
 	journal={arXiv preprint arXiv:2604.11564},
 	year={2026}
}

@article{ge2026dual,
title={Dual-Branch Remote Sensing Infrared Image Super-Resolution},
 	author={Ge, Xining and Chang, Gengjia and Yuan, Weijun and Li, Zhan and Chen, Zhanglu and Yao, Boyang and Chen, Yihang and Deng, Yifan and Liu, Shuhong},
 	journal={arXiv preprint arXiv:2604.10112},
 	year={2026}
}

@article{chang2026beyond,
title={Beyond Model Design: Data-Centric Training and Self-Ensemble for Gaussian Color Image Denoising},
author={Chang, Gengjia and Ge, Xining and Yuan, Weijun and Li, Zhan and Song, Qiurong and Zhu, Luen and Liu, Shuhong},
journal={arXiv preprint arXiv:2604.11468},
year={2026}
}

@article{ge2026clip,
title={Clip-guided data augmentation for night-time image dehazing},
 	author={Ge, Xining and Yuan, Weijun and Chang, Gengjia and Li, Xuyang and Liu, Shuhong},
 	journal={arXiv preprint arXiv:2604.05500},
 	year={2026}
}

@article{liu2026elog,
 	title={ELoG-GS: Dual-Branch Gaussian Splatting with Luminance-Guided Enhancement for Extreme Low-light 3D Reconstruction},
 	author={Liu, Yuhao and Wang, Dingju and Zheng, Ziyang},
 	journal={arXiv preprint arXiv:2604.12592},
 	year={2026}
}

@article{fu2026smokegs,
 	title={SmokeGS-R: Physics-Guided Pseudo-Clean 3DGS for Real-World Multi-View Smoke Restoration},
 	author={Fu, Xueming and Han, Lixia},
 	journal={arXiv preprint arXiv:2604.05301},
 	year={2026}
}

@article{cao2026gensmoke,
 	title={GenSmoke-GS: A Multi-Stage Method for Novel View Synthesis from Smoke-Degraded Images Using a Generative Model},
 	author={Cao, Qida and Hu, Xinyuan and Shi, Changyue and Ding, Jiajun and Yu, Zhou and Yu, Jun},
 	journal={arXiv preprint arXiv:2604.03039},
 	year={2026}
}

@article{zhu2026naka,
 	title={Naka-GS: A Bionics-inspired Dual-Branch Naka Correction and Progressive Point Pruning for Low-Light 3DGS},
 	author={Zhu, Runyu and Dong, SiXun and Zhang, Zhiqiang and Ye, Qingxia and Xu, Zhihua},
 	journal={arXiv preprint arXiv:2604.11142},
 	year={2026}
}
}

\end{document}